\documentclass[smallcondensed]{svjour3}     % onecolumn (ditto)
\smartqed  % flush right qed marks, e.g. at end of proof
\usepackage{bm}
\usepackage{graphicx}
\usepackage{mathptmx}      % use Times fonts if available on your TeX system
\usepackage{latexsym}
\usepackage{algorithm}
\usepackage{algpseudocode}
\usepackage{subfigure}
\usepackage{url}
\usepackage{comment}
\begin{document}

\title{Word Network Topic Model: A Simple but General Solution for Short and Imbalanced Texts}


\titlerunning{Word Network Topic Model}        % if too long for running head

\author{Yuan Zuo         \and
        Jichang Zhao	  \and
        Ke Xu
}

\authorrunning{Zuo et al.} % if too long for running head

\institute{Yuan Zuo \at 
			State Key Lab of Software Development Environment, Beihang University\\
			\email{skywatcher.buaa@gmail.com}
           \and
           Jichang Zhao \at
           School of Economics and Management, Beihang University\\
           \email{jichang@buaa.edu.cn}
           \and
           Ke Xu \at
           State Key Lab of Software Development Environment, Beihang University\\
           \email{kexu@nlsde.buaa.edu.cn}
}

\date{Received: date / Accepted: date}

\maketitle

\begin{abstract}
The short text has been the prevalent format for information of Internet in recent decades, especially with the development of online social media, whose millions of users generate a vast number of short messages everyday. Although sophisticated signals delivered by the short text make it a promising source for topic modeling, its extreme sparsity and imbalance brings unprecedented challenges to conventional topic models like LDA and its variants. Aiming at presenting a simple but general solution for topic modeling in short texts, we present a word co-occurrence network based model named WNTM to tackle the sparsity and imbalance simultaneously. Different from previous approaches, WNTM models the distribution over topics for each word instead of learning topics for each document, which successfully enhance the semantic density of data space without importing too much time or space complexity. Meanwhile, the rich contextual information preserved in the word-word space also guarantees its sensitivity in identifying rare topics with convincing quality. Furthermore, employing the same Gibbs sampling with LDA makes WNTM easily to be extended to various application scenarios. Extensive validations on both short and normal texts testify the outperformance of WNTM as compared to baseline methods. And finally we also demonstrate its potential in precisely discovering newly emerging topics or unexpected events in Weibo at pretty early stages.
\end{abstract}

\section{Introduction}
\label{intro}
With the rapid development of World Wide Web and the spur of various kinds of web applications, short texts have been becoming the dominating content of Internet, such as web search snippets, micro-blogs (tweets), forum messages, and news titles. Specifically, around 250 million active users in Twitter generate almost 500 million tweets everyday, which carry sophisticated signals reflecting the real world. Hence accurately mining topics behind these short texts are essential for a wide range of tasks, including content analysis~\cite{RamageEtAl:10,tong:topicdiscovery,jiang:websearchtopics}, query suggestion~\cite{jiang:querysuggestion,kais:querysuggestion}, document classification~\cite{shorttexts:classification} and text clustering~\cite{shorttexts:clustering,shorttext:similarity}. However, due to the severe sparse context information, revealing topics from short texts is still a challenging problem for traditional frameworks that initially designed to handle the normal text. Meanwhile, short texts also suffer from significantly imbalanced document distribution. For example, in social media like Weibo, the amount of entertainment tweets is much larger than the number of ones in other categories~\cite{yu2011trends,yu2013dynamics,fan2014enter}. Since the objective of most commonly used topic models is to maximize the probability of the observed data, they tend to sacrifice the performance on rare topics~\cite{imbalance:wordseeds}. Consequently, those topic models may not perform well in extrinsic tasks~\cite{topicmodel:extrinsictasks}. 

As a canonical form of existing topic models, LDA~\cite{Blei:LDA} is a hierarchical parametric Bayesian approach for topic discovery in a large corpus. To be specific, LDA models documents as mixture of topics and each topic is a probability distribution over words in the vocabulary of the corpus. Statistical inference methods are then used to learn the probability distribution over words associated with each topic and the distribution over topics for each document. Generally speaking, LDA-like models group semantically related words into a single topic by utilizing document-level word co-occurrence information~\cite{word:cooccurrence:documentlevel}, which makes them extremely sensitive to the document length and the number of documents related to each topic. Since the short text contains low word counts, those models would fail to obtain the accurate picture of how words are related to each other. Moreover, if the distribution over documents for the topic is heavily skewed, LDA-like models tend to learn more general topics held by majority documents rather than rare topics contained only by fewer documents. A recent study from Tang
et al. ~\cite{understand:lda} suggests that if the distribution over documents for each topic is heavily skewed, identifying topics from a small number of documents will be extremely difficult for LDA. While in fact rare topics might be essential for newly emerging events discovery or real-time hot trends detection in online social media~\cite{chen:twitter:topic}.

Many efforts actually have been devoted to tackle the inefficiency of LDA in modeling short texts. For example, related short texts can be aggregated into lengthy pseudo-documents before training the topic model~\cite{twitterrank:twitteraggregation} or models trained from external data (e.g. Wikipedia) can be used to help the topic inference in short texts~\cite{aggregate:topicfeature:www}. Besides, many arbitrary manipulations of LDA have also been introduced to satisfy the demands of specific short texts analyses~\cite{onetopic:eachtweeter,twitter:summarization,chen:twitter:topic}. Different from aforementioned approaches that are highly data-dependent or task-dependent, topic models focusing on the general-domain short text is emerging recently. A typical example is the biterm topic model~\cite{shorttext:btm}, which works well on short texts. However, biterm topic model is a special form of the mixture of unigram and not based on LDA. Therefore it does not overcome the shortcomings of LDA-like approaches on short texts and it's flexibility is also greatly constrained.

On the other hand, with respect to the topic imbalance, performance improvement of LDA is mainly obtained by adding prior information to guide the topic learning progress~\cite{imbalance:wordseeds,mustlink:mustnotlink} or using asymmetric Dirichlet prior over the document-topic distribution~\cite{rethinkinglda:priors}. While note that in practice, the knowledge about underlying structure of a given corpus is often undiscovered, so the prior information is not easily acquirable. As to asymmetric Dirichlet priors, how to determine proper parameter estimations is sophisticated and scenario-dependent for different applications and assuming symmetric Dirichlet priors help most variants of LDA keep the flexibility. Therefore, alleviating topic imbalance of LDA with symmetric Dirichlet priors is actually quite desirable.

To sum up, approaches mentioned above are neither scenario independent nor easy to be extended. In order to handle the sparsity and imbalance of short texts simultaneously through a general framework, we propose \emph{Word Network Topic Model} (WNTM) based on the word co-occurrence network. The main idea of WNTM comes from the following observations. 1) When texts are short, word-by-document space is extremely sparse, while word-word space is still rather dense. Since the topic quality can be guaranteed in the dense word-word space~\cite{arora:practicaltopicmodeling}, we conjecture that learning topic components from word co-occurrence network rather than document collection is more reliable. 2)  Intuitively, the number of words connected to rare topics often exceeds the amount of documents related to those topics. So the distribution over topics for words is greatly less skewed than the distribution over topics for documents. 3) Since the distribution over topics for each document can not be learned accurately in short or imbalanced texts, we should learn the distribution over topics for each word instead. 4) Different from the existing solutions, a new framework should be simple enough to guarantee its scalability in different application scenarios. Hence WNTM employs the standard Gibbs sampling for LDA to discover latent word group (i.e. topics)~\cite{lda:groupdiscovery} and learns distribution over topics for words rather than topics for documents. Learning word's topics rather than document's topics makes WNTM less sensitive to the document length or heterogeneity of the topic distribution. In addition, the word co-occurrence network can be constructed with any type of given texts, which makes WNTM further simple and general in real-world applications. 

Extensive experiments are conducted on various data sets to compare WNTM and baseline methods in three aspects, including topic quality, word semantic similarity and document semantic classification. And results suggest that WNTM can discover the most coherent topics in short texts. Meanwhile, WNTM outperforms all baseline methods on word similarity and document categorization in both short and normal texts. Particularly, WNTM shows much better capability than LDA in rare topic detection in extremely imbalanced texts. Major contributions of this paper are 1) WNTM is a generative model for a word network rather than a collection of documents, 2) and it learns topics for each word rather than topics for each document, therefore 3) it is less sensitive to document length or document distribution over topics. 4) Since WNTM uses the standard Gibbs sampling for LDA, it's general and very easy to be applied in different scenarios. 

The rest of the paper is organized as follows. We first give a short review of relative works in section~\ref{sec:relatedwork}. This is followed by detail introductions of our model and re-weighting method in section~\ref{sec:wntm} and section~\ref{sec:reweigh}. Experimental results are illustrated and explained in section~\ref{sec:exp}. Finally, we conclude the present work briefly in section~\ref{sec:conclusion} and several possible directions in future are also pointed out. 
\section{Related works}
\label{sec:relatedwork}
Probabilistic topic models such as PLSA~\cite{Hofmann:PLSA} and LDA~\cite{Blei:LDA} have been extensively applied in exploring text corpora. Particularly, LDA is a more complete generative model since it extends PLSA by adding Dirichlet priors on topic distributions. Due to their extensibility, many complicated variants of LDA and PLSA have been proposed in the last decade, such as the dynamic topic model~\cite{dynamic:topicmodel}, social topic model~\cite{social:topicmodel}, author-topic model~\cite{author:topicmodel} and author-topic-community model~\cite{authortopic:community} etc. While most of them are designed to handle normal texts with special additional properties, such as time, social relationship and authorship.

The sparse short texts has also attract much research interest in the previous literature and most early studies mainly focus on increasing data density through utilizing auxiliary information. For example, Hong et al.~\cite{Empiricaltopicmodel:twitter} train topic models on aggregated tweets that sharing the same word, and find those models work better than those being directly trained on original tweets. Sahami et al.~\cite{Sahami:2006} propose a search-snippet-based similarity measure for short texts. Jin et al. learn topics on short texts via transfer learning from auxiliary long text data~\cite{shorttexts:clustering}. Another way to deal with data sparsity in short texts is to apply special topic models. For example, Zhao et al. assume each tweet only covers a single topic~\cite{onetopic:eachtweeter}. Yan et al.~\cite{shorttext:btm} propose a special form of mixture of unigrams~\cite{topicmodel:mixtureofunigram}, which is called biterm topic model to improve topic modeling on short texts.

While regarding to the topic modeling on imbalanced texts, the prior knowledge has been widely used to alleviate skewed distributions over documents of different topics. Andrzejewski et al. propose Dirichlet forest priors to incorporate must-links and cannot-links constraints into topic models~\cite{mustlink:mustnotlink}. Chen et al. use general lexical knowledge to help discovering coherent topics~\cite{general:priorknowledge}. It is worthy noting that the document level knowledge can also be utilized. For example, Ramage et al. bring labels into the generative process of the corpora~\cite{labeled:topicmodel,multilabel:topicmodel} and Blei et al propose a supervised topic model~\cite{supervised:topicmodel} to predict the category labels for input labeled documents.

However, different from above approaches, we try to figure out a simple but general solution to take care of sparsity and imbalance in texts simultaneously. To the best of our knowledge, little attention has been paid to this issue and our proposed topic model is the first one handling short and imbalanced texts in a general way without exploiting any external knowledge.

\section{Word network topic model}
\label{sec:wntm}
Commonly used topic models implicitly take advantage of rich word co-occurrence patterns in documents. However, short texts naturally lack of enough contextual information. Furthermore, the goal of traditional topic models is to maximize the probability of generating observed documents, and rare topics that reflected by fewer documents are tended to be ignored. As a result, directly applying conventional topic models on short or imbalanced texts can not perform as well as that on normal balanced texts. In order to solve the problem mentioned above through a simple but general method, we propose a new framework, which applies the same Gibbs sampling~\cite{stand:gibbsampling} with LDA to discover latent word groups in a word co-occurrence network. Here latent word groups of the network are taken as topic components of a corpus. In addition, the distribution over latent groups for each word is also learned by our model. The details of the new framework is presented in the following subsections.
\subsection{Word co-occurrence network}
\label{subsec:wcn}
In a word co-occurrence network (which may also be denoted as \emph{word network} in the following text if there is no conflict), nodes are words occurring in the corpus and an edge between two words indicates that the connected two words have co-occurred in the same context at least once. Here the context can refer to a document or a sliding window with fixed size. To limit the size of word network and reserve only the local context for each word, we take a sliding window of fixed size as the context in the present work. We empirically set the size of siding window to 10 (a typical value employed in the previous study) in this paper, since a word is only semantically related with adjacent words, especially in the short text. Degree of a node is defined as the sum of weights over its adjacent links. While activity of a node is defined as the averaged weight of its adjacent links.

In order to convert the given document collection into a word network, we first filter out stopping words and low frequency words, and then a sliding window is moved to scan each document. As the window scanning word by word through the document, any two distinct words appear in the same window would be regarded as co-occurred with each other. Times that two words co-occurred are accumulated and defined as the weight of the corresponding edge between them.

Note that in topic models, a topic can be viewed as a bag of words co-occurred frequently in the same document, which is very similar to latent word groups (or communities) in the word network, since words co-occurred frequently in the same sliding windows are closely connected in the semantic space and they could appear in the same document with high probabilities. Therefore, we could take latent word groups in our word network based model as the topics in LDA. At the same time, learning topics from word co-occurrence network, a special form of word-word space, has a theoretical guarantee for topic coherence according to the work of Arora et al.~\cite{arora:practicaltopicmodeling}. What's more, rare topics may form compact latent word groups in word network, therefore a topic model based on the word network could effectively find word groups that correspond to rare topics. Based on the considerations above, we propose our \emph{word network topic model} (WNTM). In order to keep the new model simple and general to employ at different scenarios, we take a similar approach as Keith et al. did in~\cite{lda:groupdiscovery} to discover latent word groups in word network.

\subsection{Word network topic model (WNTM)}
\label{subsec:wntm}

\begin{figure}
\centering
\includegraphics[width=\textwidth]{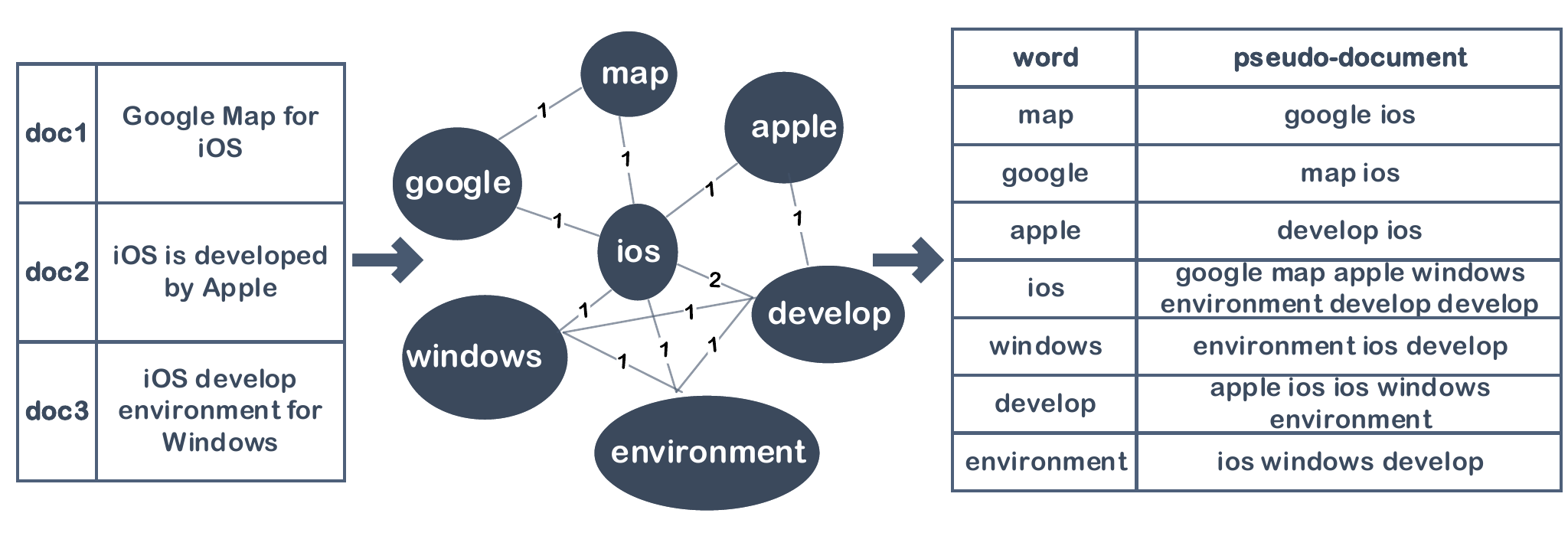}
\caption{An illustration of the input for WNTM.}
\label{fig:framework}
\end{figure}

The standard Gibbs sampling for LDA can be used to discover latent word groups in large word network. While in order to reserve the standard Gibbs sampling, we first have to represent the word co-occurrence network back to a pseudo-document set. We assume the word network as undirected and weighted. As illustrated in Fig.~\ref{fig:framework}, each word in the network can be treated as a pseudo-document with content constituted by the list of its adjacent words. Obviously since the word network is weighted, the adjacent words may occur multiple times in the text of the pseudo-document.

Although WNTM uses the same Gibbs Sampling with LDA, the rationalities underlying the generative process of them are different. LDA learns to generate a collection of documents by using topics and words under those topics. However, WNTM learns to generate each word's adjacent word-list in the network by using latent word groups and words belonging to those groups. More specifically, WNTM learns the statistical relations between words, latent word groups and words' adjacent word-lists by assuming that each word's adjacent word-list is generated semantically by a particular probabilistic model. It first supposes that there is a fixed set of latent word groups in the word network, and each latent word group $z$ is associated with a multinomial distribution over the vocabulary $\Phi_z$, which is drawn from a Dirichlet prior $Dir(\beta)$. The generative process of the whole pseudo-document collection converted from the word network can be interpreted as follows:
\begin{enumerate}
\item For each latent word group $z$, draw $\Phi_z \sim Dir(\beta)$, a multinomial distribution over words for $z$
\item Draw $\Theta_i \sim Dir(\alpha)$, a latent word group distribution for the adjacent word-list $L_i$ of the word $w_i$
\item For each word $w_j\in L_i$:
 \begin{enumerate}
  \itemsep0em 
  \item select a latent word group $z_j \sim \Theta_i$
  \item select the adjacent word $w_j \sim \Phi_{z_j}$
 \end{enumerate}
\end{enumerate}
In WNTM, the $\Theta$ distributions represent the probability of latent word groups appearing in each word's adjacent word-list and the $\Phi$ distributions stand for the probability of words belonging to each latent word group. Given the observed corpus, WNTM first converts it to a word network, then generate the pseudo-document set and finally the same Gibbs sampling implementation that developed for conventional LDA is employed to infer values of the latent variable in both $\Phi$ and $\Theta$. Because each word's adjacent word-list actually represents its global context information, so different from previous LDA-like approaches, WNTM models the distribution over latent word groups for each word instead of the distribution over topics for each document.

\subsection{Inferring topics in a document}
\label{subsec:otd}
As discussed in the previous section, WNTM does not model the document generation process. Therefore, we cannot directly obtain topics in a document from the result of Gibbs sampling. Since WNTM models the generation process of each word's adjacent word-list which stands for the word's global contextual information, we can take topic proportions of word $w_i$'s adjacent word-list $\Theta_i$ as topic proportions in $w_i$. Given topic proportions for all words, topics of each document can be obtained accordingly. Specifically, to infer topics in a document, we assume that the expectation of the topic proportions of words generated by a document equals to the topic proportions of the document, i.e.,
\begin{equation}\abovedisplayskip\belowdisplayskip
P(z|d)=\sum_{w_i}P(z|w_i)P(w_i|d),
\end{equation}
where $P(z|w_i)$ equals to $\Theta_{i,z}$, which has been learned in WNTM. As to $P(w_i|d)$, we simply take the empirical distribution of words in the document as a estimation, i.e.,
\begin{equation}\abovedisplayskip\belowdisplayskip
P(w_i|d)=\frac{n_d(w_i)}{Len(d)},
\end{equation}
where $n_d(w_i)$ is the word frequency of $w_i$ in document $d$ and $Len(d)$ is the length of $d$. It is worthy noting the above strategy is straight-forward and easy to implement, which guarantees WNTM's simplicity further.

To sum up, when texts are short and sparse, learning topics in word-by-document space will suffer from the severe sparsity problem, while learning topics in a word-word space has a theoretical guarantee for topic coherence, which has been proved in~\cite{arora:practicaltopicmodeling}. Meanwhile, as the distribution over documents for each topic is imbalanced, rare topics tend to be ignored by LDA-like models. However, we conjecture that words related to rare topics would still form a semantically compact latent group in the word co-occurrence network. So latent groups standing for rare topics could also be detected by WNTM. Therefore, the rich contextual information in word-word space facilitates WNTM to discover topics in word co-occurrence network other than directly reveal topics from document collection. 

\section{Complexity analysis and word network re-weighting}
\label{sec:reweigh}
Although our inference on WNTM uses the same Gibbs sampling with LDA, the running time and space complexities of them are different. We will compare the time complexity in detail first, and then give a brief discussion about the space complexity of the two models. Finally, In order to reduce the time and space complexity of WNTM, we further propose a method to perform word network re-weighting.

\subsection{Complexity analyses}
\label{subsec:ca}
The time complexity of LDA is $O(N_d K_z L_d )$, where $N_d$ is the number of documents, $K_z$ is the number of topics and $L_d$ is the average document length. Similarly, the time complexity for WNTM is $O(N_p K_g L_p )$, where $N_p$ is the number of pseudo-documents, i.e., the size of the vocabulary, $K_g$ is the number of latent word groups (topics) and $L_p$ is the average pseudo-document length. Since the maximum number of sliding windows in a corpus is $N_dL_d$, and each sliding window can generate ${c \choose 2}$ edges, where $c$ is the size of the sliding window. Thus, approximately $N_p L_p$ can be rewritten as
\begin{equation}\abovedisplayskip\belowdisplayskip
\label{eq:tc}
N_p L_p \approx N_d L_d c(c-1).
\end{equation}
Supposing $K_z$ equals to $K_g$, the time complexity of WNTM is $o(c^2)$ times larger than LDA's cost. In practice, for short texts, the average document length $\langle l\rangle$ is often small than $c$. So when applied to short texts, the time complexity of WNTM is acceptable. However, with respect to normal texts, it becomes unacceptable since $c$ can be set to a large number. 

The space complexity of LDA is $O(N_d K_z+N_d L_d)$ and the space complexity of WNTM is $O(N_p K_g+N_p L_p)$. Similar to the time complexity, if we assume $N_d$ is equal to $N_p$ and $K_z$ is equal to $K_g$ , then the memory WNTM consumes is $o(c^2)$ times of the size that LDA needs. In order to reduce the time (or space) complexity, for instance, to decrease the time (or space) complexity to linear times the cost of LDA, we would propose a word network re-weighting method in the next subsection, which could help boost the learning process of WNTM effectively. 

\subsection{Word network re-weighting}
\label{subsec:rew}
The above analysis shows that the time and space complexity is unaffordable for Gibbs sampling over the pseudo-document collection, which is directly generated by the weighted word network. To reduce both the time and space consumption, we need to decrease $N_p L_p$. While $N_p$ is fixed as it is determined by the size of the given corpus's vocabulary, so $L_p$ is the only tunable parameter. Representing the length of a pseudo-document, $L_p$ also equals to the degree of the node corresponding to the pseudo-document. Therefore, decreasing the weights of edges in the network can shorten the length of pseudo-documents, and then reduce the time and space complexity of WNTM accordingly. In order to reserve the relative closeness of different words in the process of tuning weights, a re-weighting method is illustrated in Algorithm~\ref{alg:rew}. 
\begin{algorithm}[h]
    \caption{Word network re-weighting algorithm}
    \label{alg:rew}
    \begin{algorithmic}[1]
       \Require
               the original word network $G=(V,E,W)$, where $V$ is the set of words, $E$ is the set of edges and $W$ is the set of weights for edges.
       \Ensure
              the re-weighted word network $G'=(V,E,W').$
       \Statex
       \State compute degree $D(n)$ and activity $A(n)$ of each node $n\in V$
       \ForAll{$e=(n_1,n_2) \in E$}
       \State set $w_e=\left\lceil \frac{w_e}{A(n_i)}\right\rceil$, $\mathop{argmin}\limits_{i}\{D(n_i),~i=1,2\}$
       \EndFor
   \end{algorithmic}
\end{algorithm}

The weight of each edge is divided by the activity of its end with lower degree, denoted as $w_e.$ Along this line, we can decrease the whole weighted degree of the entire word network, which can then decrease the time and space complexity of WNTM. Since the weighted degree of a node must be larger than $c-1$, then the averaged weighted degree of the word network is actually much larger than $c-1.$ Therefore, the averaged length of pseudo-documents should be smaller than $\frac{L_p}{c-1}$, where $L_p$ is the averaged length before re-weighting. Then from Eq.~\ref{eq:tc} we can easily get that the time and space cost of re-weighted WNTM is $O(c)$ times the complexity of LDA, which is just a linear scale-up. Hence the re-weighting algorithm successfully reduces the cost of WNTM and guarantees its feasibility in both short and normal texts. 

\section{Experiments}
\label{sec:exp}
In this paper, we evaluate our approach in three measures, including topic quality, word similarity and document classification. For each measure, extensive experiments are performed on real-world short texts and normal texts respectively. For short texts, We take LDA and biterm topic model (BTM)~\cite{shorttext:btm} as baseline methods. As to normal texts, we omit the comparison with BTM because of its intense time complexity when applied on normal texts. what is worth mentioning is that the comparison between WNTM and LDA can indicate the strengths and weaknesses of learning topics from document collection and word co-occurrence network.

Most experiments in this section are carried out on a Windows Server with an Intel Xeon 2.40GHz CPU and 12G memory except for the experiments using Wikipedia data set. Due to the large volume of the Wikipedia data set, corresponding experiments are conducted on a Linux cluster with 13 nodes. Each node contains 2 Intel Xeon 2.27Hz CPUs and 12 GB memory. For both LDA and WNTM, we use a java open-source implementation JGibbLDA\footnote{\url{http://jgibblda.sourceforge.net/}} on short texts and an MPI open-source implementation titled PLDA\footnote{\url{http://code.google.com/p/plda/}} on normal texts. For BTM, we use the source code opened by the authors\footnote{\url{http://code.google.com/p/btm/}}. For JGibbLDA and BTM, we set $\alpha=50/k$ and $\beta=0.01$, where $k$ is the number of topics. For PLDA, we set $\alpha=0.1$ and $\beta=0.01.$ In all experiments, number of topics is set to 100, length of the sliding window for WNTM is set to 10 and each model's Gibbs sampling is run for 2,000 iterations. Except for document classification on news contents, the results reported here are the average over 10 rounds.

\subsection{Evaluation of the topic quality}
\label{subsec:etq}
It's a typical way for evaluating topic models through comparing the perplexity on a held-out test set. However, WNTM does not model the generation process of documents. Hence, the perplexity is not suitable in this paper. Furthermore, recent research shows that the perplexity does not always correlate with semantically interpretable topics~\cite{topicmodel:extrinsictasks}. Therefore, here we utilize the topic coherence as an evaluation metric for the topic discovery, which has been found to correlate well with human judgments of the topic quality.

\subsubsection{Topic coherence}
\label{ssubsec:tc}
Topic coherence (also called UMass measure~\cite{umass:topiccoherence}) is a comprehensive and automated evaluation measure for topic models, which measures the score of a single topic by computing the semantic similarity degree between high probability words in the topic. Higher topic coherence often indicates better topic quality, i.e., better topic interpret-ability. The topic coherence is defined as
\begin{equation}
C(z;M^{(z)})=\sum_{t=2}^{T}\sum_{l=1}^{t-1}\log\frac{D(m_{t}^{(z)},m_{l}^{(z)})+\epsilon}{D(m_{l}^{(z)})}⁡,
\end{equation}
where $M^{(z)}=(m_{1}^{(z)},...,m_{T}^{(z)})$ is the list of the $T$ most probable words in topic $z$, $D(m)$ counts the number of documents containing the word $m$, $D(m ,m^{\prime})$ counts the number of documents containing both $m$ and $m^{\prime}$, and $\epsilon=10^{-12}$ is used to avoid taking the log of zero for words that never co-occur and to smooth the score for completely unrelated words. We use the average coherence score of all topics as the evaluation metric for topic quality of different topic models.

\subsubsection{Topic coherence on short texts}
\label{ssubsec:tcst}
To investigate WNTM's ability of learning high quality topics from real-world short texts, we carry out experiments on one day's micro-blogs\footnote{Publicly available at \url{http://ipv6.nlsde.buaa.edu.cn/zhaojichang/paper/wntm.rar}} sampled from Weibo. As a Twitter-like service in China, it also imposes a limited length for each tweet, i.e., no more than 140 Chinese characters. Since the textual content of micro-blogs is not formal, careful preprocessing is quite necessary. In the preprocessing, we take the following steps to wash the collected corpus: (a) using NLPIR\footnote{\url{http://ictclas.nlpir.org/downloads}} to do tokenization; (b) removing stopping words; (c) removing words with frequency less than 20; (d) filtering out URLs and non-Chinese characters; (e) removing micro-blogs with length less than 10. Finally, 189,223 micro-blogs retained with 20,942 distinct words in total. The average number of tokens in documents is 17.2. 

\begin{table}
\centering
\caption{The average topic coherence results on the micro-blog collection. A larger value stands for more coherent topics.}
\label{tab:coherence_st}
\begin{tabular}{llll}
\hline\noalign{\smallskip}
$T$ & 5 & 10 & 20 \\
\noalign{\smallskip}\hline\noalign{\smallskip}
LDA & $-$36.6$\pm$1.8 & $-$221.4$\pm$4.0 & $-$1484.6$\pm$39.4 \\
BTM & $-$37.4$\pm$1.9 & $-$207.5$\pm$8.2 & $-$1235.9$\pm$30.7 \\
\bf{WNTM} & $-$\bf{32.5}$\pm$\bf{1.3} & $-$\bf{181.6}$\pm$\bf{5.5} & $-$\bf{1056.6}$\pm$\bf{22.8} \\
\noalign{\smallskip}\hline
\end{tabular}
\end{table}

We compare WNTM with LDA and BTM on this micro-blog collection. For all models, we set the number of topics to 100. The average topic coherence of three models is listed in Table~\ref{tab:coherence_st}, where the size of top words set in each topic, denoted as $T$, ranges from 5 to 20. We find that the average topic coherence of WNTM is obviously higher than other two models, which indicates that WNTM outperforms LDA and BTM in learning high quality topics from short texts. And the improvement made by WNTM is statistically significant ($p$-value $<0.001$ by $t$-test). The outperformance of WNTM as compared to LDA is in accordance with our understanding that learning topics from dense word-word space can guarantee the topic quality even in short texts. BTM also outperforms LDA since it directly model word pairs rather than documents to solve data sparsity in short texts. Remarkably, WNTM outperforms BTM significantly. WNTM is based on the framework of LDA and BTM is special form mixture of unigram, which might be the reason why WNTM works better than BTM on short texts. What's more, we also notice that as $T$ grows, the performance gap increases and WNTM is more stable by possessing much lower deviations.

\subsubsection{Topic coherence on normal texts}
\label{ssubsec:tcnt}

We conduct experiments on Wikipedia data provided by Phan et al.~\cite{aggregate:topicfeature:www} to investigate WNTM's ability of learning high quality topics from real-world normal texts. The Wikipedia data set contains 71,986 documents with 60,649 distinct words. The average number of tokens in documents is 423.5. The number of topics is set to 100 for each model and the average topic coherence result is listed in Table~\ref{tab:coherence_nt}, where the size of top words set $T$ in each topic ranges from 5 to 20. As shown in Table~\ref{tab:coherence_nt}, the average topic coherence of WNTM is slightly higher than that of LDA when $T=5$ and 10, while slightly lower than that of LDA when $T=20$. From the results, we can see that learning topics by grouping words co-occurred in a small range of context can benefit the top 10 words' coherence in each topic, but when $T=20$, the coherence of top words might need more plentiful document-level word co-occurrence information to maintain. However, with the increasing of distance between two words, the relation between them becomes less relevant. Because of this, the difference between WNTM and LDA is not likely to be obvious. According to the results of $t$-test, two models gain no statistically significant improvement than each other. Therefore, we can conclude that WNTM can produce similar high-quality topics as LDA dose on normal texts. Note that the word-by-document space has no sparsity problem in normal texts, so LDA can utilize the rich contextual information in each document to learn high quality topics. Thus, for normal texts, learning topics from word-by-document space and word-word space makes little difference in topic quality.

\begin{table}
\centering
\caption{The average topic coherence of LDA and WNTM on the Wikipedia data set. A larger value means more coherent topics.}
\label{tab:coherence_nt}
\begin{tabular}{llll}
\hline\noalign{\smallskip}
$T$ & 5 & 10 & 20 \\
\noalign{\smallskip}\hline\noalign{\smallskip}
LDA & $-$13.9$\pm$0.3 & $-$69.7$\pm$1.0 & $-$327.4$\pm$3.3 \\
WNTM & $-$13.8$\pm$0.2 & $-$69.5$\pm$0.6 & $-$329.7$\pm$2.4\\
\noalign{\smallskip}\hline
\end{tabular}
\end{table}

\subsection{Word similarity tasks}
\label{subsec:wst}
The average topic coherence is an intrinsic measure used to evaluate the quality of all topics. Higher topic coherence often indicates better topic quality, but it does not guarantee a better performance on extrinsic tasks. The study of Keith et al.~\cite{lda:groupdiscovery} revealed that LDA is better than LSA (Latent Semantic Analysis)~\cite{lsa:1990} at learning descriptive topics, while LSA is better than LDA at creating a compact semantic representation of words and documents and outperforms LDA in extrinsic tasks~\cite{exploring:topiccoherences}. Hence, we carry out experiments to compare the performance of WNTM and other methods on extrinsic tasks such as word similarity tasks and document classification, which would help further illustrate the two models' effectiveness in creating compact semantic representation of words and documents. In this section, we compare two models' ability of learning semantic representation of words on short and normal texts, respectively. To begin with, we will introduce how to calculate the semantical similarity between two words.

\subsubsection{Semantic representation of a word}
\label{ssubsec:srw}
For LDA and BTM, the conditional topic distribution for a word $w$ can be defined as its semantic representation
$$s_w=[p(z_1|w),p(z_2|w),...,p(z_k|w)],$$
where $k$ is the number of topics. $p(z_k|w)$ is easy to obtain after Gibbs sampling stage is completed,
\begin{equation}
p(z_k|w)=\frac{n_{w|z_k}}{n_w},
\end{equation}
where $n_{w|z_k}$ stands for how many times $w$ has be assigned with topic $k$ during the sampling and $n_w$ means the total occurrence of $w$ in given corpus. With respect to WNTM, we do not have to calculate $p(z_k|w)$. The $\theta_w$ can be directly used as the semantic representation of word $w$, since WNTM learns the topic distribution over words as matrix $\Theta$. Therefore, we use $\Theta$'s row vector $\theta_w$ corresponding to $w$ as its semantic representation, which is denoted as
$$s_w=\theta_w=[\Theta_{w,1},\Theta_{w,2},...,\Theta_{w,k}].$$
Then we can measure the distance of two words by use of the Jensen-Shannon divergence
\begin{equation}
JS(s_i,s_j)=\frac{1}{2}D_{kl}(s_j \parallel m)+\frac{1}{2}D_{kl}(s_i \parallel m),
\end{equation}
where $s_i$ and $s_j$ are the semantic representations of words $i$ and $j$, $m=\frac{1}{2}(s_i+s_j)$ and $D_{kl}(p||q)=\sum_ip_i\ln\frac{p_i}{q_i}$ is the Kullback-Leibler divergence. If we consider the topic distributions as space vectors, cosine similarity can also be used to measure the distance of two words. It can be defined as
\begin{equation}
Cosine(s_i,s_j)=\frac{s_i\cdot s_j}{\parallel s_i\parallel \parallel s_j\parallel}.
\end{equation}

\subsubsection{Word similarity tasks}
\label{ssubsec:wst}
Word similarity tasks are widely used to evaluate distributional semantic spaces. Topics learned by topic models can be viewed as the knowledge about semantic distributions of words. If a topic model learns topics accurately, then we can expect similar words, such as ``man'' and ``woman'', to be represented with similar semantic representations.
In this paper, we use the word similarity task presented by Wang et al.~\cite{wordsim:chinese} to evaluate the ability of word semantic modeling of two models on Weibo data set. Regarding to normal texts, we use word similarity tasks designed by Rubenstein and Goodenough in~\cite{wordsim:353:1} and Finkelstein et al. in~\cite{wordsim:353:2} on Wikipedia data set. In each task, the semantic similarity between given pairs of words were evaluated by human. The word similarity task introduced by Wang et al. contains 240 pairs of Chinese words and each pair's semantic relatedness is rated from 0-10, in which a higher score reflects a more semantically similar word pair. The task introduced by Finkelstein was constituted by 353 pairs of English words and Rubenstein and Goodenough's task contains 65 English words. Each pair was also given a human rate indicating the pair's semantic closeness.

\begin{figure}
\centering
\includegraphics[width=0.5\textwidth]{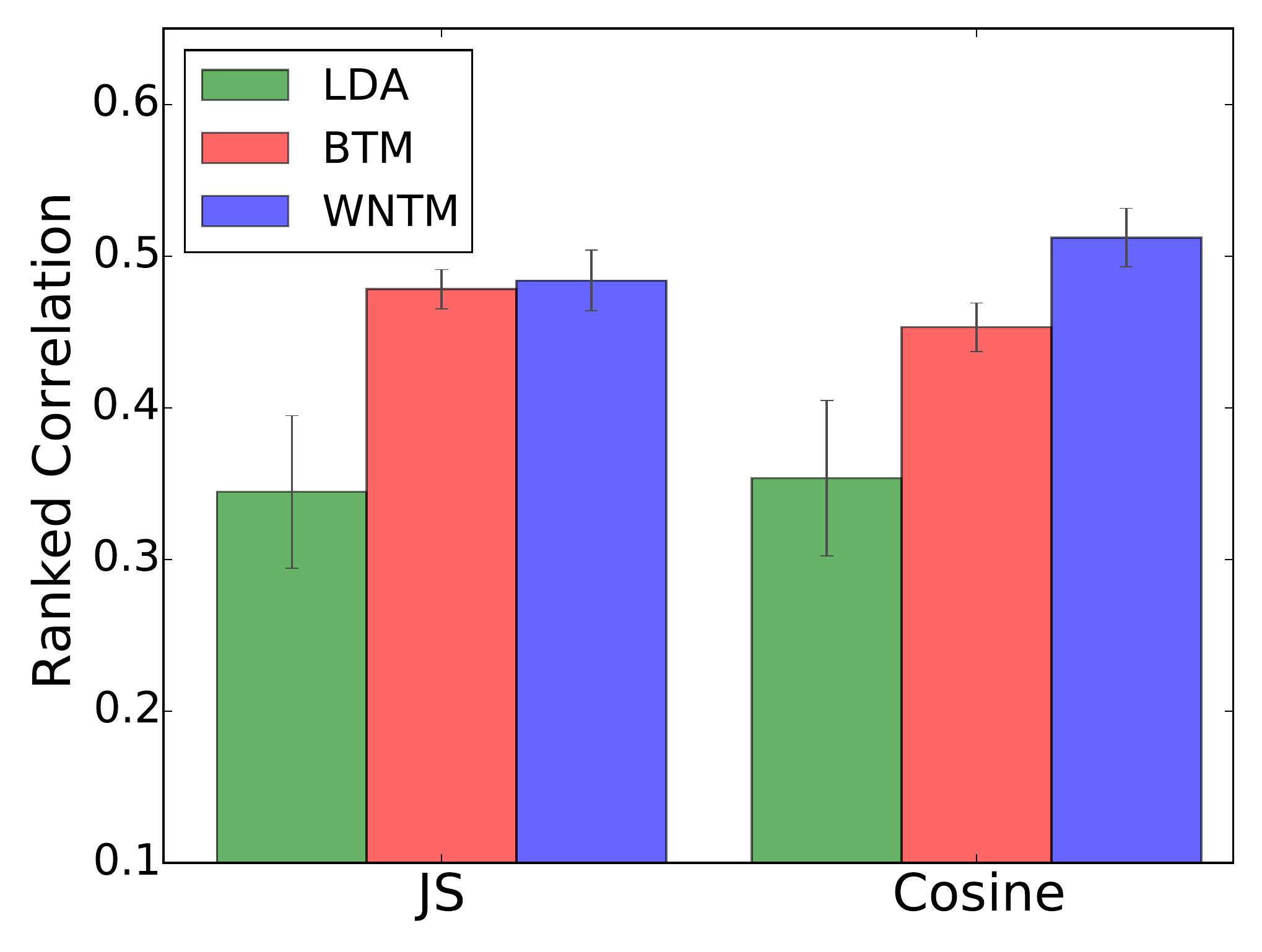}
\caption{Ranked correlation results of word similarity task on Weibo data. The result demonstrates that WNTM outperforms LDA and BTM on creating better semantic representations of words on short texts.}
\label{fig:weibo_ws}
\end{figure}

We evaluate topic models by calculating the similarity between each pair of words through two similarity measurements (JS is short for Jensen-Shannon divergence and Cosine is short for Cosine similarity) in the evaluate set and then compare the model's ratings with human ratings by the ranked correlation. Intuitively higher correlation indicates better word semantic modeling. The number of topics is set to 100 for both models and the ranked correlation result on Weibo data is illustrated in Fig.~\ref{fig:weibo_ws}. 

From the result, we can see that both WNTM and BTM outperforms LDA significantly, no matter using JS as the similarity measure or Cosine. WNTM performs similar with BTM on JS, but outperforms BTM significantly on Cosine. Through directly modeling word pairs other than documents, BTM successfully avoid the document-level data sparsity issue. WNTM also directly model word co-occurrences to alleviate document-level data sparsity. Therefore, BTM and WNTM can learn more accurate word semantic representations than LDA. However, WNTM apply the framework of LDA to model word's context extracted from word network, therefore, the model assumption of WNTM is more accordance with actual data circumstance as compared to BTM. Thus although both WNTM and BTM surpass LDA in learning word semantic representations, WNTM is more reliable than BTM.

\begin{figure} \centering   
\subfigure[Word similarity task by Finkelstein et al.] { \label{subfig:wordsim_353}    
\includegraphics[width=0.45\columnwidth]{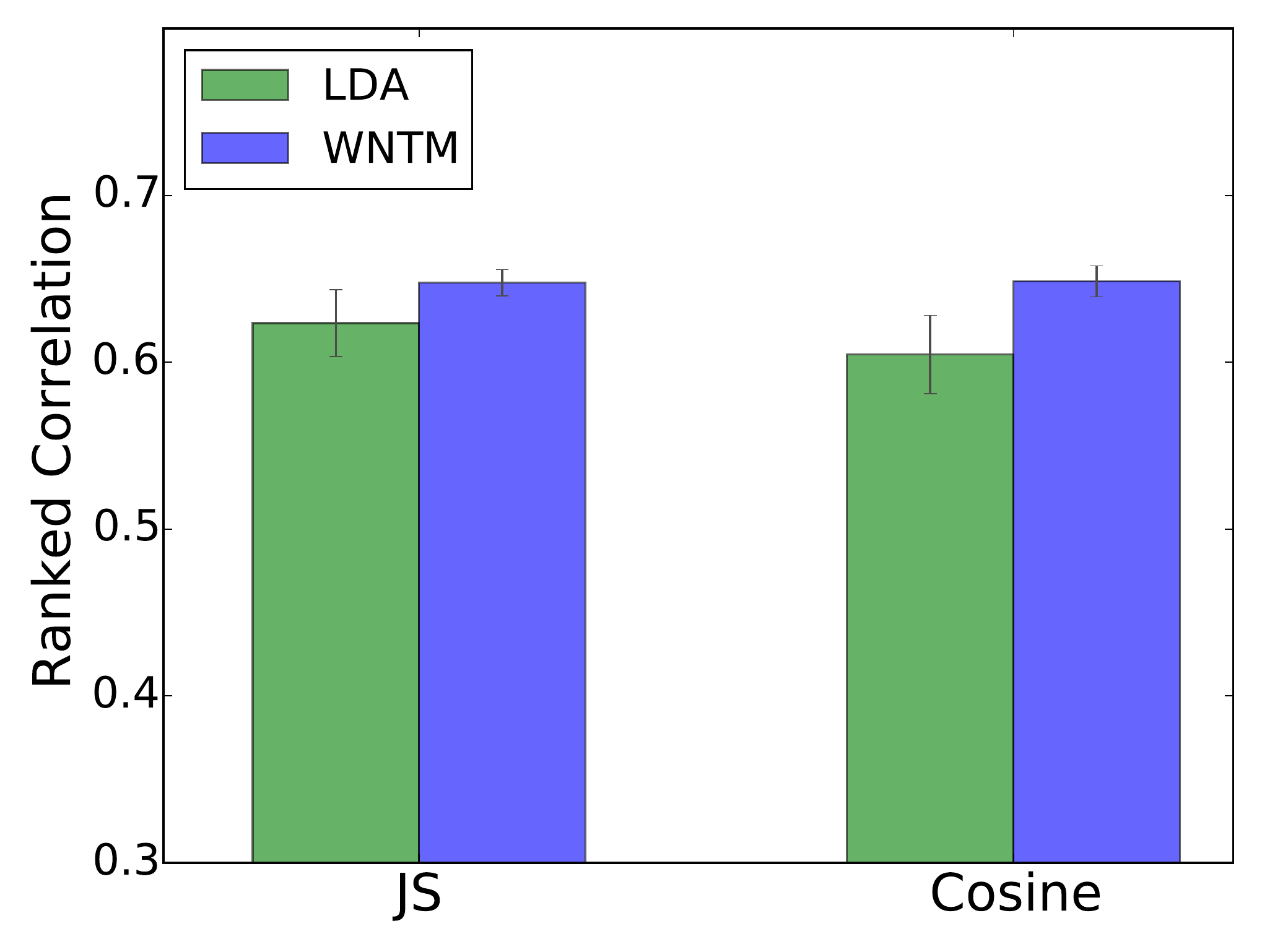} 
}    
\subfigure[Word similarity task by Rubenstein et al.] { \label{subfig:wordsim_65}    
\includegraphics[width=0.45\columnwidth]{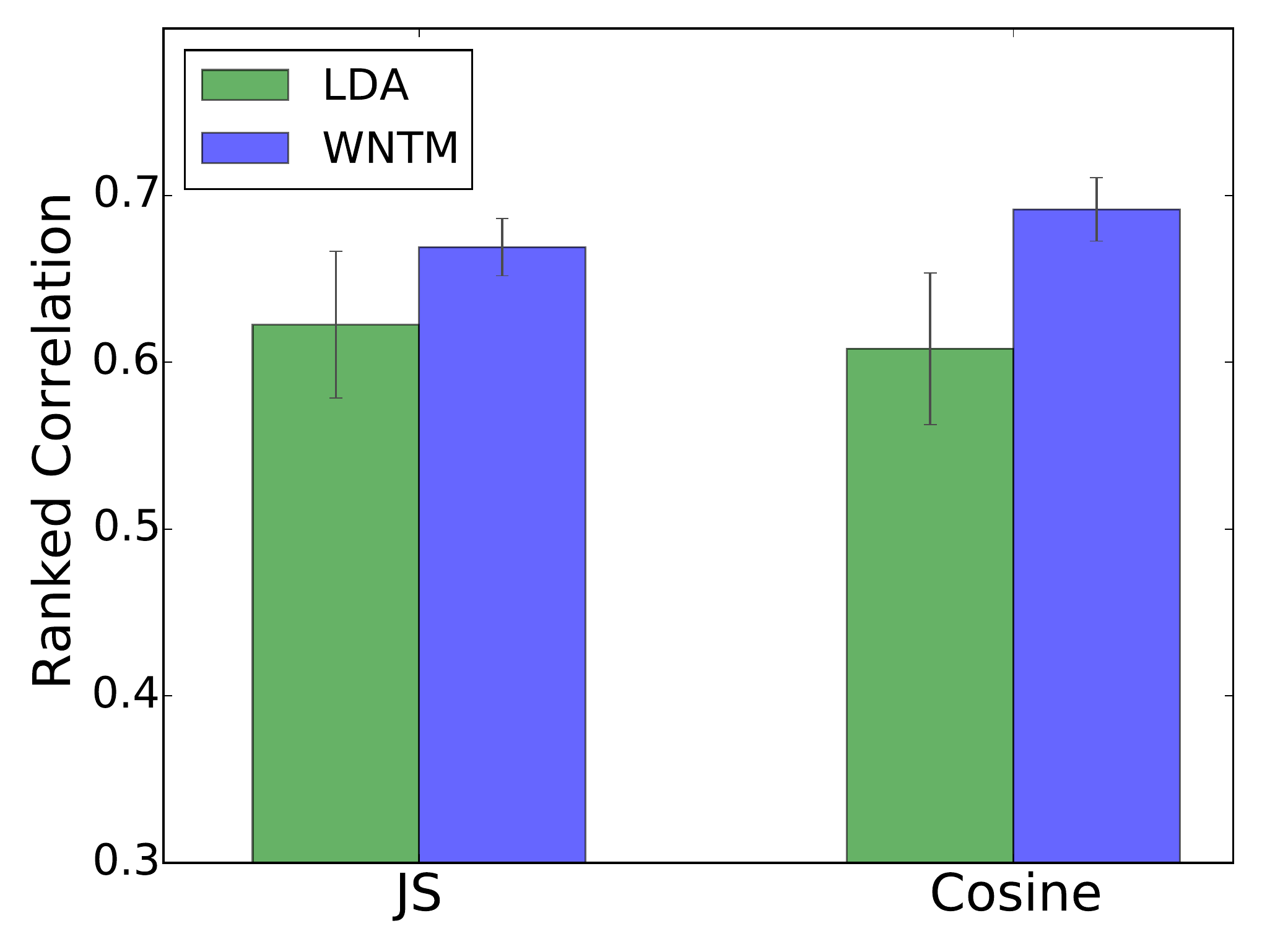}    
}    
\caption{Ranked correlation results of word similarity task on Wikipedia data set. The result shows that WNTM even outperforms LDA on creating better semantic representations of words on normal texts.}    
\label{fig:wiki_ws}    
\end{figure}

The result of Wikipedia data is illustrated in Fig.~\ref{fig:wiki_ws}. We surprisingly find that WNTM still outperforms LDA on both similarity measurements. The performance gap between two models on normal texts shrinks compared to the result on short texts, since the document-level data sparsity problem is gone. However, WNTM's result is more stable than LDA by possessing much lower deviations. Although sharing similar topic coherence results in normal texts, the two models' ability in creating compact semantic representation for words are different and WNTM performs better than LDA in word similarity tasks.

\subsection{Document classification}
\label{subsec:dc}
In order to compare these models' ability in learning semantic representation of documents on short and normal texts, we evaluate them by performing document classifications on news titles and news content respectively in this section. To illustrate WNTM's advantage on imbalanced texts as compared to LDA, we further conduct classification experiments by tuning the heterogeneity of topic distributions.

\begin{table}
\centering
\caption{Label counts for news titles and contents used in document classification tasks. Here Ent represents Entertainment.}
\label{tab:dc_nt}
\subtable[Label counts for news titles]{
\begin{tabular}{llll}
\hline\noalign{\smallskip}
Label & Count & Label & Count \\
\noalign{\smallskip}\hline\noalign{\smallskip}
Finance & 31414 & Car & 6532 \\
Sports & 25414 & IT & 2321 \\
Society & 14889 & Military & 1733 \\
Ent & 11208 & House & 1410 \\
Lady & 8128 & Culture & 983 \\
Olympics & 7117 & Health & 962 \\
Education & 6594 && \\
\noalign{\smallskip}\hline
\end{tabular}
}
\subtable[Label counts for news content]{
\begin{tabular}{llll}
\hline\noalign{\smallskip}
Label & Count & Label & Count \\
\noalign{\smallskip}\hline\noalign{\smallskip}
Finance & 133480 & Car & 18675 \\
Sports & 115946 & IT & 10650 \\
Society & 70743 & Military & 8706 \\
Ent & 53335 & House & 6407 \\
Olympics & 34767 & Health & 2340 \\
Lady & 31689 & Culture & 2334 \\
Education & 19482 & & \\
\noalign{\smallskip}\hline
\end{tabular}
}
\end{table}

\subsubsection{Evaluation on news corpus}
\label{ssubsec:enc}
To explore the effectiveness of WNTM in document semantic modeling on short texts and normal texts, we first evaluate its performance on document classification tasks of news titles and news content, which are extracted from the news corpus provided by Sogou.com\footnote{\url{http://www.sogou.com/labs/dl/ca.html}}. After preprocessing, we obtain 508,554 news titles with label distributions listed in Table~\ref{tab:dc_nt}, and 59,348 distinct words in total. The average number of token in news titles is 5.5. We also obtain 118,705 news reports with label distributions listed in Table~\ref{tab:dc_nt}, and 76,114 distinct words in total. The average number of token in each report is 175.9.

\begin{figure}
\centering
\subfigure[News titles]{\includegraphics[width=0.45\textwidth]{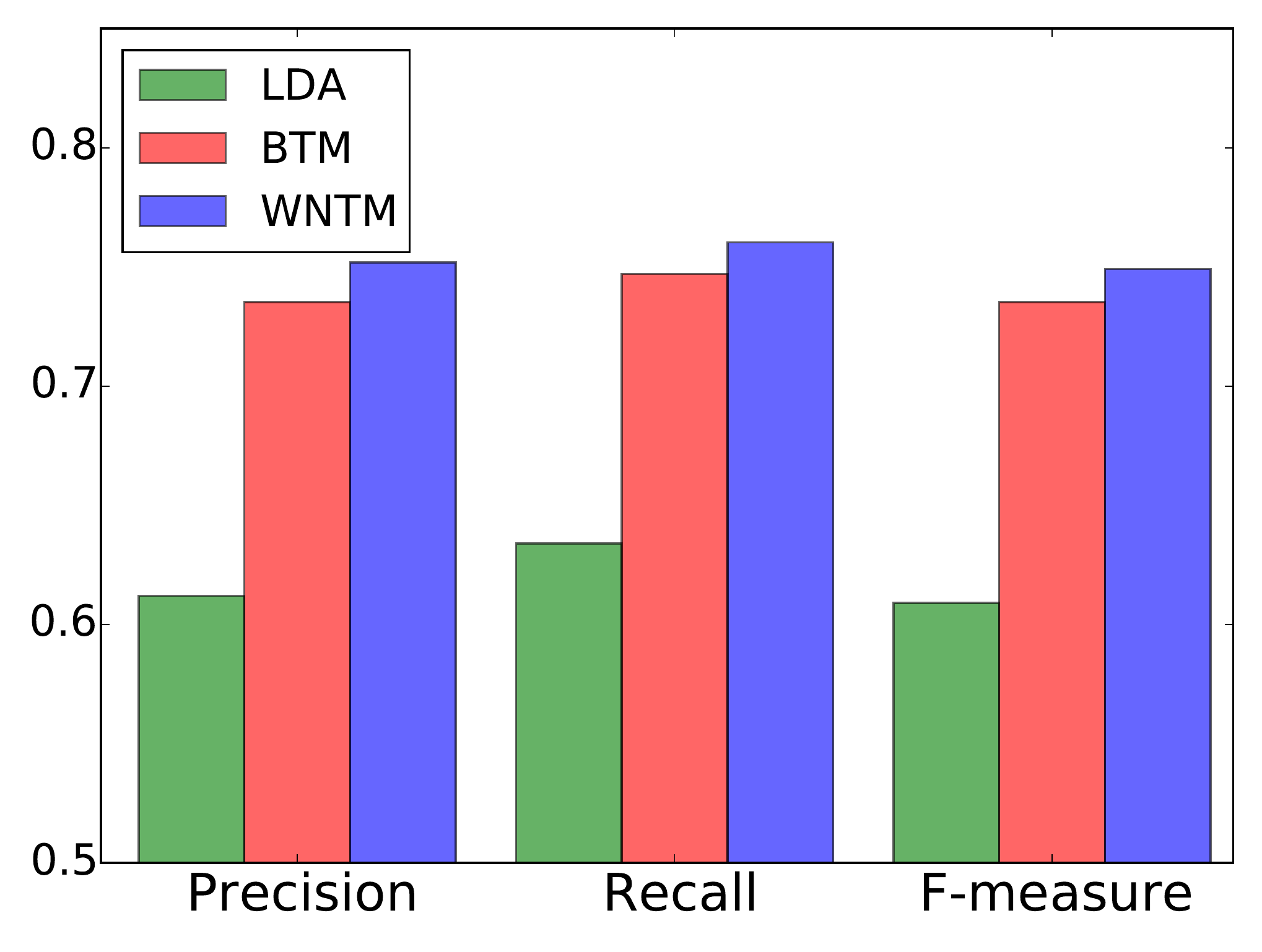}
\label{fig:dc_nt}}
\subfigure[News content]{\includegraphics[width=0.45\textwidth]{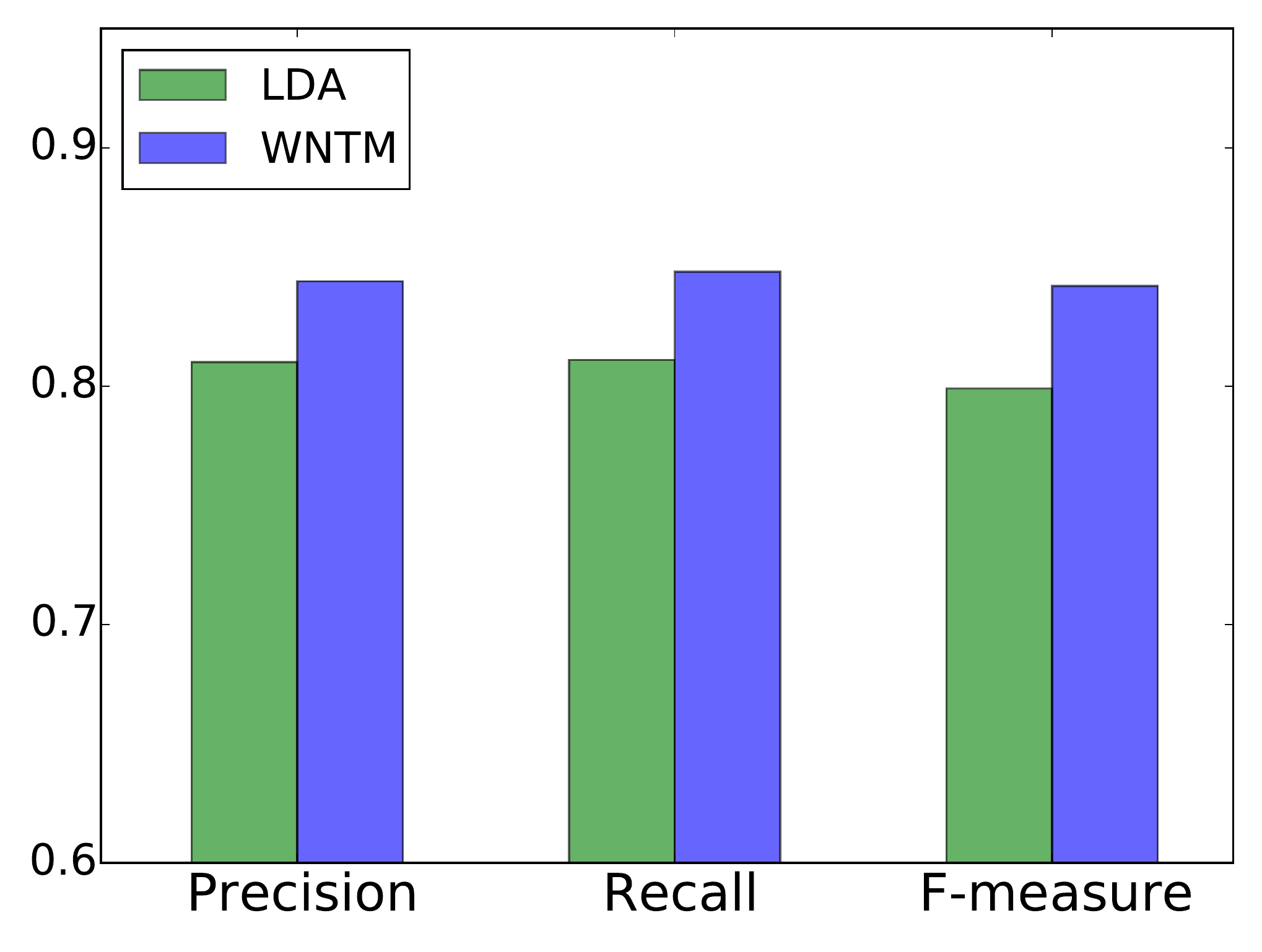}
\label{fig:dc_nc}}
\caption{Classification performance comparison between WNTM and LDA on short and normal texts, respectively.}
\end{figure}

Taking topic model as a method of dimensionality reduction, we can reduce a document into a fixed set of topics, which can be features for document categorization. For each topic model trained on 100 topics, we perform 10-fold cross-validation on news titles (or content). In each fold, we randomly split news titles (or content) into training and test subsets with the ratio 9:1, and classified them by using LIBLINEAR\footnote{\url{http://www.csie.ntu.edu.tw/~cjlin/liblinear/}}. The weighted average accuracy (precision), recall and F-measure for news titles are shown in Fig.~\ref{fig:dc_nt}. The result for news content is shown in Fig.~\ref{fig:dc_nc}. 

From the result, we can find that WNTM and BTM outperforms LDA in classifying short texts and normal texts. The divergence in the performance of classifying short texts suggests that the data sparsity problem seriously affects LDA, while less variation is found in WNTM and BTM. WNTM performs better than BTM, similar to word similarity task. Regarding to normal texts, LDA's classification result becomes acceptable since the data is no longer sparse. However, WNTM still outperforms LDA, which indicates that WNTM is better than LDA in recognizing the resemblance of documents even for normal texts.

\begin{figure} \centering   
\subfigure[LDA] { \label{subfig:label_lda}    
\includegraphics[width=0.7\columnwidth]{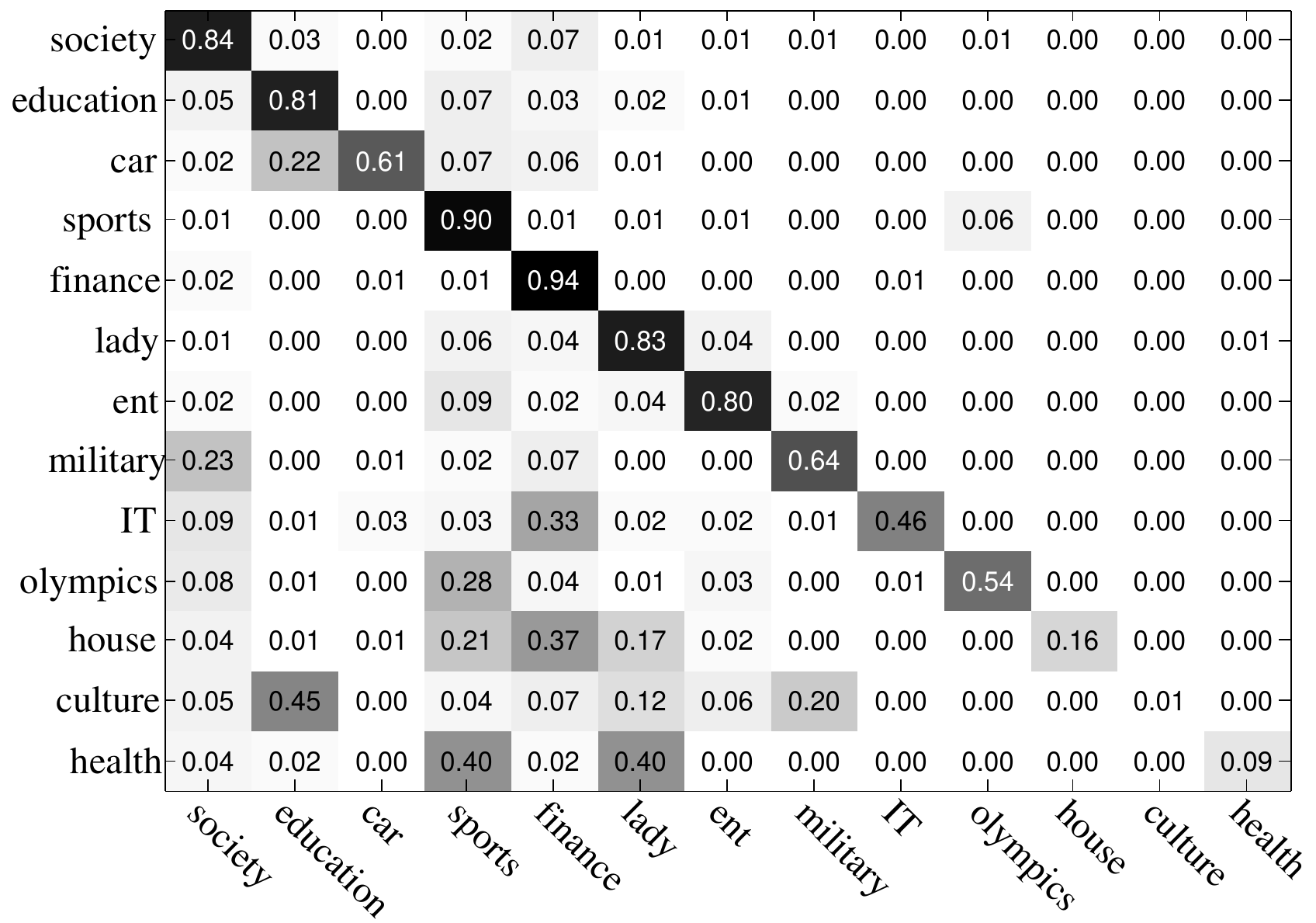} 
}    
\subfigure[WNTM] { \label{subfig:label_wntm}    
\includegraphics[width=0.7\columnwidth]{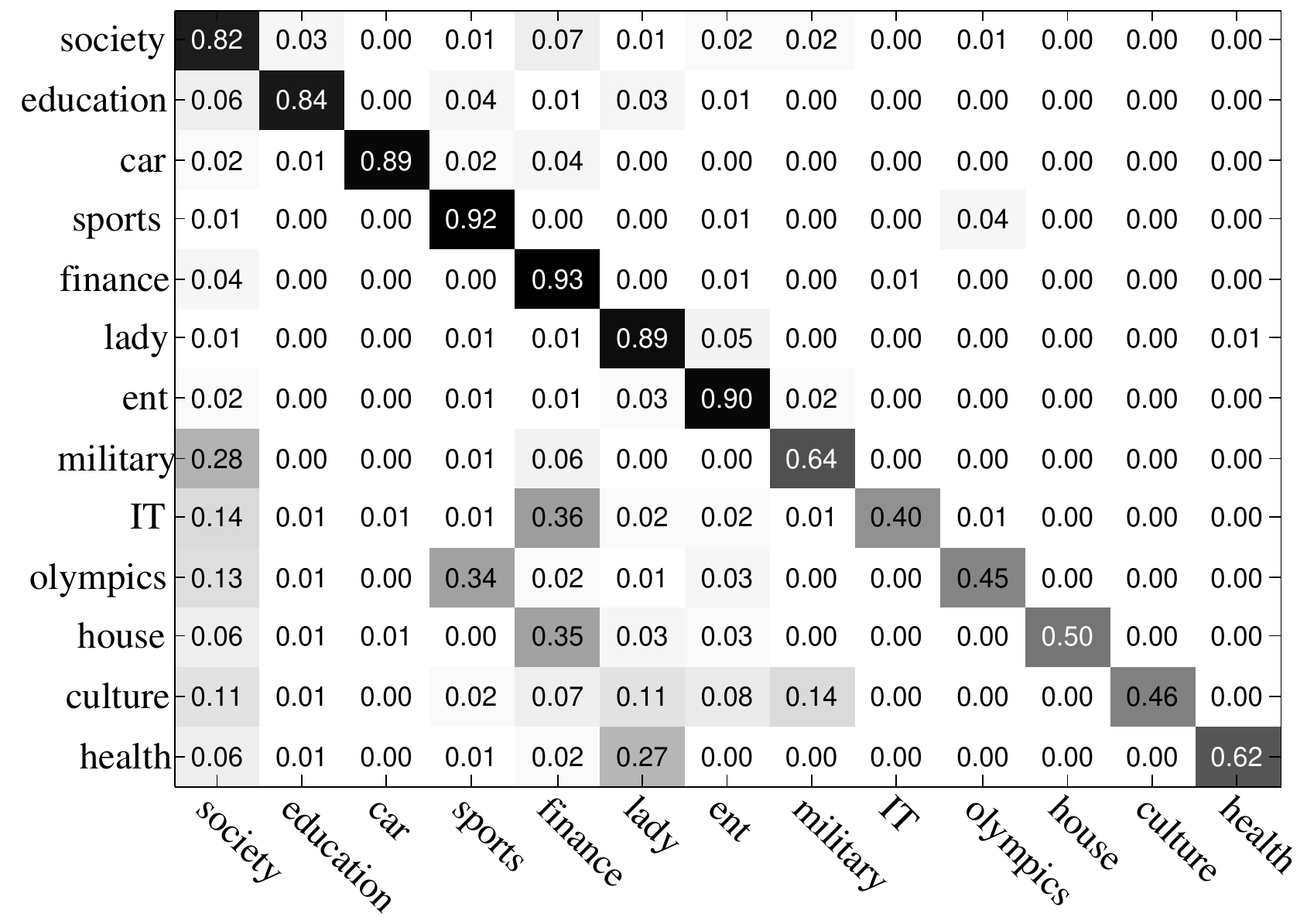}    
}    
\caption{Confusion matrix for news content classification. The result from news title is similar and hence it is not reported here.}    
\label{fig:label_confusion}    
\end{figure}

Fig.~\ref{fig:label_confusion} shows the confusion matrix of LDA and WNTM in the news contents classification. For example, LDA confuses ``House'' with ``Sports'', ``Finance'' and ``Lady'', where ``House'' and ``Finance'' may be hard to distinguish because they are strongly relevant, while obviously ``House'' has few connections with ``Sports'' and ``Lady''. Consequently, LDA almost misclassifies all documents under label ``House''. The similar situation also happens to ``Culture'' and ``Health''. As listed in Table~\ref{tab:dc_nt}, the significant confusion of those labels mainly comes from their low numbers of documents and in fact they represent exactly the rare topics in the employed corpus. However, as shown in Fig.~\ref{subfig:label_wntm}, WNTM still achieves a relative better performance as compared to LDA. It indicates that WNTM gains outstanding improvement on the performance of identifying rare labels, which evidently justifies the conjecture that WNTM can identify more rare topics than LDA as the text is imbalanced.

\subsubsection{Document classification on imbalanced texts}
\label{ssubsec:dcit}
\begin{figure}
\centering
\includegraphics[width=0.65\textwidth]{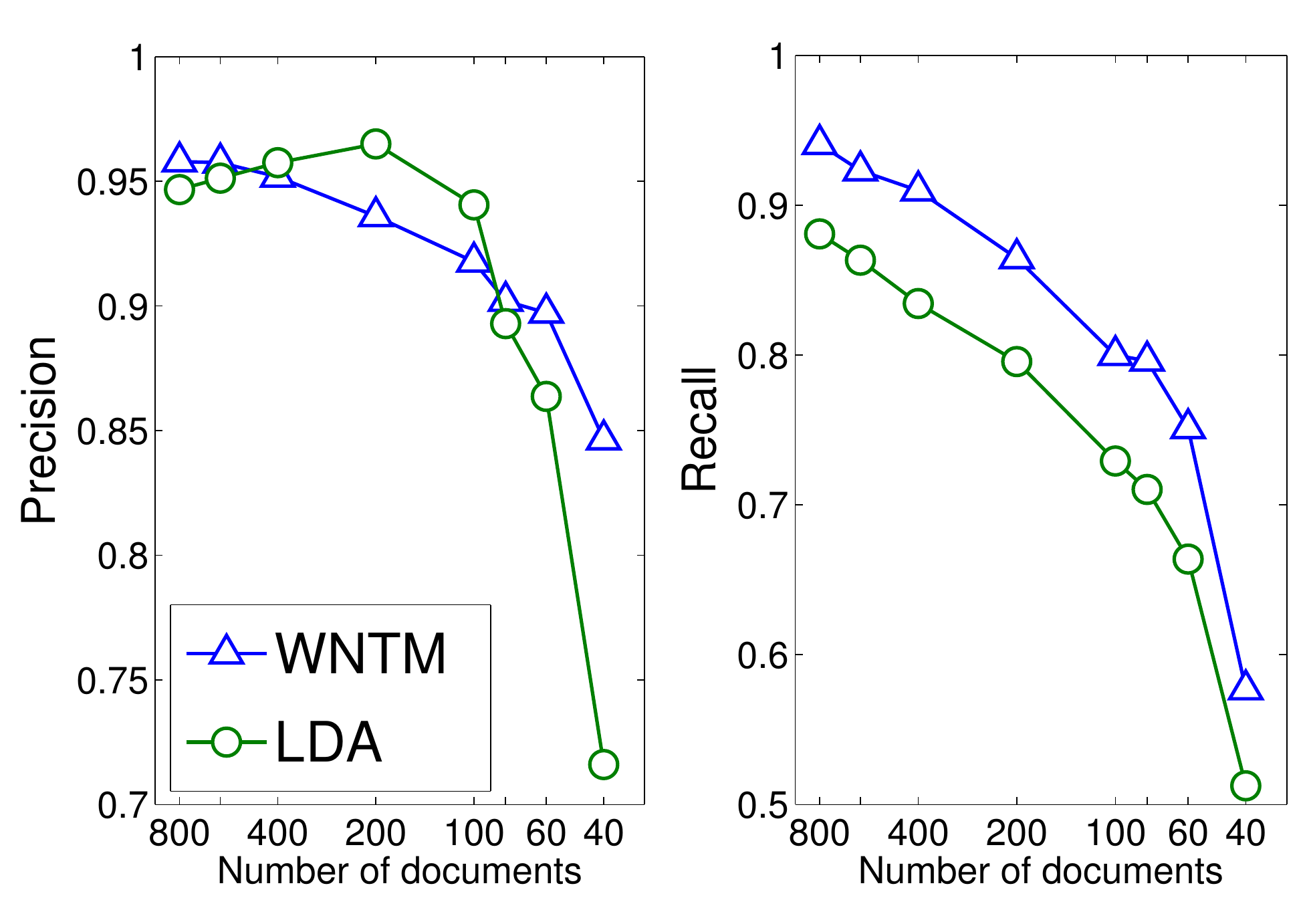}
\caption{Average precision and recall on classifying ``Car'' news contents in the imbalanced document classification. The x-axis is set in log scale.}
\label{fig:dc_car}
\end{figure}

To compare performance between WNTM and LDA in classifying documents containing rare topics further, we investigate the variation of their classification results by continuously tuning the imbalance of texts. We first build a balanced data set from news content introduced previously. This balanced data set includes news from five classes, namely ``Education'', ``Car'', ``Finance'', ``Lady'' and ``Ent''. Within each class, there are 1,000 documents allocated equally. Second, we take various numbers of documents away from ``Car'', which is randomly selected, to build data sets with different levels of imbalance. Specifically, we build 8 groups of imbalanced data sets, each group contains 1,000 documents belong to classes except ``Car'', which contains $d_c$ documents, where $d_c$ is tunable parameter and we let $d_c=800,~600,~400,~200,~100,~80,~60$ and 40. As $d_c$ ranges from 800 to 40, the imbalance of data set is enhanced accordingly. To avoid the influence of the randomness in taking away documents from ``Car'', we randomly sample documents under ``Car'' 15 times for each group. The average precision and recall on classifying ``Car'' documents is illustrated in Fig.~\ref{fig:dc_car}. From the result, we can find that the average precision of LDA first slightly increases when the number of news under ``Car'' reducing from 800 to 200, and then decreases dramatically due to the more and more serious imbalance of the data set. On the contrary, the average precision of WNTM decreases slowly with the enhancement of the imbalance. Particularly, when the number of ``Car'' documents equals to 60 and 40, the average precision of LDA has decreased to 86\% and 72\% while WNTM's average precision is 90\% and 85\%. This result indicates that WNTM is more accurate than LDA in distinguishing documents under rare labels. Note that as $d_c$ varies in the range between 200 and 500, the averaged precision of LDA is trivially greater than that of WNTM, which is caused by fluctuations and would not affect our above analysis. From results of average recall, we can find that WNTM and LDA have similar trending with the imbalance enhancement, while WNTM always outperforms LDA in terms of the average recall.  To sum up, we can conclude that WNTM performs better than LDA in dealing with text classification on imbalanced data sets.

From the above experiments, we find WNTM dominates LDA in learning rare topics. Thus it is reasonable to conjecture that WNTM could be a better choice than LDA in detecting newly emerging topics or unexpected events at early stages in social media, like Twitter or Weibo. To illustrate this point, we collect 10,000 micro-blogs from Weibo, and then inject $d_m$ micro-blogs related to the event of ``MH370'' into them, where $d_m$ ranges from 10 to 100. Note that the 10,000 micro-blogs have no relation with ``MH370'' since they are posted exactly before the event happens. For different settings of $d_m$, we train both models 10 times since the result of Gibbs sampling varies each time. After training, we look over the top 20 words of each topic to search the topic word ``MH370'', and count how many times it is found for each setting of $d_m$. From the results we find that WNTM can always identify a topic with word ``MH370'' contained by its top 20 word-list when $d_m\ge 30$. However, LDA achieves the same result only as $d_m\ge 50.$ Thus, WNTM is more sensitive to rare topics and can learn newly emerging topics at earlier stages than LDA. Besides, we also evaluate the quality of the topic related to ``MH370'' by human review. If a topic's top words contains more ``M7370'' related words than others, we evaluate this topic as the best. We list top 15 words of the best ``MH370'' topic learned by two models for comparison when $d_m=50$ and 100. It is interesting that when $d_m=50,$ the best ``MH370'' topic learned by LDA contains 8 words that never appear in the 50 injected ``MH370'' micro-blogs, while WNTM only has 4. When $d_m=100$, LDA's ``MH370'' topic still contains 4 unrelated words, while WNTM only has 1 error word left. Based on these results, we can conclude that WNTM can detect event topic earlier than LDA and the quality of the topic is much better than that identified by LDA.

\section{Conclusions}
\label{sec:conclusion}
A simple but general approach named WNTM is presented in this paper to facilitate the topic modeling in short and imbalanced texts at acceptable cost. Different from conventional LDA-like solutions, it explores topics from word co-occurrence networks and successfully alleviates the data sparsity and the topic-document heterogeneity in word-by-document space. Thorough experiments on both short and normal texts suggest that WNTM outperforms baseline methods in tasks of topic coherence, word similarity and document classification. Furthermore, the ability of capturing rare topics with high quality indicates that WNTM could be an effective model for detecting newly emerging topics or unexpected events in social media at quite early stages. However, promising new research directions still exist for further study. For example, we would like to investigate the influence imposed on topic results as different means of establishing word co-occurrence networks are taken. Besides, we may also consider of using a word's neighbor words within a given semantic distance to model its topics.

\begin{acknowledgements}
This work was supported by 863 Program (Grant No. 2012AA011005) and Research Fund for the Doctoral Program of Higher Education of China (Grant No. 20111102110019). Jichang Zhao was partially supported by 863 Program (Grant No. 2014AA015203) and the Fundamental Research Funds for the Central Universities (Grant Nos. YWF-14-RSC-109 and YWF-14-JGXY-001).
\end{acknowledgements}

\bibliographystyle{spmpsci}      % mathematics and physical sciences
\bibliography{refs}   % name your BibTeX data base

\begin{thebibliography}{10}
\providecommand{\url}[1]{{#1}}
\providecommand{\urlprefix}{URL }
\expandafter\ifx\csname urlstyle\endcsname\relax
  \providecommand{\doi}[1]{DOI~\discretionary{}{}{}#1}\else
  \providecommand{\doi}{DOI~\discretionary{}{}{}\begingroup
  \urlstyle{rm}\Url}\fi

\bibitem{mustlink:mustnotlink}
Andrzejewski, D., Zhu, X., Craven, M.: Incorporating domain knowledge into
  topic modeling via dirichlet forest priors.
\newblock In: ICML, pp. 25--32 (2009)

\bibitem{arora:practicaltopicmodeling}
Arora, S., Ge, R., Halpern, Y., Mimno, D.M., Moitra, A., Sontag, D., Wu, Y.,
  Zhu, M.: A practical algorithm for topic modeling with provable guarantees.
\newblock In: ICML, vol. 28 (2), pp. 280--288. JMLR: W\&CP (2013)

\bibitem{dynamic:topicmodel}
Blei, D.M., Lafferty, J.D.: Dynamic topic models.
\newblock In: ICML, pp. 113--120 (2006)

\bibitem{Blei:LDA}
Blei, D.M., Ng, A.Y., Jordan, M.I.: Latent dirichlet allocation.
\newblock J. Mach. Learn. Res. \textbf{3}, 993--1022 (2003)

\bibitem{social:topicmodel}
Cha, Y., Cho, J.: Social-network analysis using topic models.
\newblock In: SIGIR, pp. 565--574 (2012)

\bibitem{topicmodel:extrinsictasks}
Chang, J., Gerrish, S., Wang, C., Boyd-graber, J.L., Blei, D.M.: Reading tea
  leaves: How humans interpret topic models.
\newblock In: Advances in Neural Information Processing Systems 22, pp.
  288--296 (2009)

\bibitem{shorttexts:classification}
Chen, M., Jin, X., Shen, D.: Short text classification improved by learning
  multi-granularity topics.
\newblock In: IJCAI, pp. 1776--1781 (2011)

\bibitem{chen:twitter:topic}
Chen, Y., Amiri, H., Li, Z., Chua, T.S.: Emerging topic detection for
  organizations from microblogs.
\newblock In: SIGIR, pp. 43--52 (2013)

\bibitem{general:priorknowledge}
Chen, Z., Mukherjee, A., Liu, B., Hsu, M., Castellanos, M., Ghosh, R.:
  Discovering coherent topics using general knowledge.
\newblock In: CIKM, pp. 209--218 (2013)

\bibitem{twitter:summarization}
Chua, F.C.T., Asur, S.: Automatic summarization of events from social media.
\newblock In: ICWSM (2013)

\bibitem{lsa:1990}
Deerwester, S., Dumais, S.T., Furnas, G.W., Landauer, T.K., Harshman, R.:
  Indexing by latent semantic analysis.
\newblock Journal of the American Society for Information Science
  \textbf{41}(6), 391--407 (1990)

\bibitem{fan2014enter}
Fan, R., Zhao, J., Feng, X., Xu, K.: Topic dynamics in weibo: Happy
  entertainment dominates but angry finance is more periodic.
\newblock In: Conference on Advances in Social Networks Analysis and Mining
  (ASONAM), 2014 IEEE/ACM International, pp. 230--233 (2014)

\bibitem{wordsim:353:2}
Finkelstein, L., Gabrilovich, E., Matias, Y., Rivlin, E., Solan, Z., Wolfman,
  G., Ruppin, E.: Placing search in context: The concept revisited.
\newblock ACM Trans. Inf. Syst. \textbf{20}(1), 116--131 (2002)

\bibitem{stand:gibbsampling}
Heinrich, G.: {Parameter estimation for text analysis}.
\newblock Web:http://www.arbylon.net/publications/text-est.pdf  (2005)

\bibitem{lda:groupdiscovery}
Henderson, K., Eliassi-Rad, T.: Applying latent dirichlet allocation to group
  discovery in large graphs.
\newblock In: SAC, pp. 1456--1461 (2009)

\bibitem{Hofmann:PLSA}
Hofmann, T.: Probabilistic latent semantic indexing.
\newblock In: SIGIR, pp. 50--57 (1999)

\bibitem{Empiricaltopicmodel:twitter}
Hong, L., Davison, B.D.: Empirical study of topic modeling in twitter.
\newblock In: SOMA, pp. 80--88 (2010)

\bibitem{imbalance:wordseeds}
Jagarlamudi, J., Daum{\'e} III, H., Udupa, R.: Incorporating lexical priors
  into topic models.
\newblock In: EACL, pp. 204--213 (2012)

\bibitem{jiang:querysuggestion}
Jiang, D., Leung, K.T., Vosecky, J., Ng, W.: Personalized query suggestion with
  diversity awareness.
\newblock In: ICDE, pp. 400--411 (2014)

\bibitem{jiang:websearchtopics}
Jiang, D., Leung, K.W.T., Ng, W.: Fast topic discovery from web search streams.
\newblock In: Proceedings of the 23rd International Conference on World Wide
  Web, WWW, pp. 949--960. ACM, New York, NY, USA (2014)

\bibitem{shorttexts:clustering}
Jin, O., Liu, N.N., Zhao, K., Yu, Y., Yang, Q.: Transferring topical knowledge
  from auxiliary long texts for short text clustering.
\newblock In: CIKM, pp. 775--784 (2011)

\bibitem{authortopic:community}
Li, C., Cheung, W., Ye, Y., Zhang, X., Chu, D., Li, X.: The
  author-topic-community model for author interest profiling and community
  discovery.
\newblock Knowledge and Information Systems pp. 1--25 (2014)

\bibitem{supervised:topicmodel}
Mcauliffe, J.D., Blei, D.M.: Supervised topic models.
\newblock In: J.~Platt, D.~Koller, Y.~Singer, S.~Roweis (eds.) Advances in
  Neural Information Processing Systems 20, pp. 121--128 (2008)

\bibitem{rethinkinglda:priors}
McCallum, A., Mimno, D.M., Wallach, H.M.: Rethinking lda: Why priors matter.
\newblock In: Y.~Bengio, D.~Schuurmans, J.~Lafferty, C.~Williams, A.~Culotta
  (eds.) NIPS, pp. 1973--1981. Curran Associates, Inc. (2009)

\bibitem{umass:topiccoherence}
Mimno, D., Wallach, H.M., Talley, E., Leenders, M., McCallum, A.: Optimizing
  semantic coherence in topic models.
\newblock In: EMNLP, pp. 262--272 (2011)

\bibitem{topicmodel:mixtureofunigram}
Nigam, K., McCallum, A.K., Thrun, S., Mitchell, T.: Text classification from
  labeled and unlabeled documents using em.
\newblock Mach. Learn. \textbf{39}(2-3), 103--134 (2000)

\bibitem{aggregate:topicfeature:www}
Phan, X.H., Nguyen, L.M., Horiguchi, S.: Learning to classify short and sparse
  text \& web with hidden topics from large-scale data collections.
\newblock In: WWW, pp. 91--100 (2008)

\bibitem{shorttext:similarity}
Quan, X., Liu, G., Lu, Z., Ni, X., Wenyin, L.: Short text similarity based on
  probabilistic topics.
\newblock Knowledge and Information Systems \textbf{25}(3), 473--491 (2010)

\bibitem{RamageEtAl:10}
Ramage, D., Dumais, S., Liebling, D.: Characterizing microblogs with topic
  models.
\newblock In: ICWSM (2010)

\bibitem{labeled:topicmodel}
Ramage, D., Hall, D., Nallapati, R., Manning, C.D.: Labeled lda: A supervised
  topic model for credit attribution in multi-labeled corpora.
\newblock In: EMNLP, pp. 248--256 (2009)

\bibitem{author:topicmodel}
Rosen-Zvi, M., Griffiths, T., Steyvers, M., Smyth, P.: The author-topic model
  for authors and documents.
\newblock In: UAI, pp. 487--494 (2004)

\bibitem{wordsim:353:1}
Rubenstein, H., Goodenough, J.B.: Contextual correlates of synonymy.
\newblock Commun. ACM \textbf{8}(10), 627--633 (1965)

\bibitem{multilabel:topicmodel}
Rubin, T., Chambers, A., Smyth, P., Steyvers, M.: Statistical topic models for
  multi-label document classification.
\newblock Machine Learning \textbf{88}(1-2), 157--208 (2012)

\bibitem{Sahami:2006}
Sahami, M., Heilman, T.D.: A web-based kernel function for measuring the
  similarity of short text snippets.
\newblock In: WWW, pp. 377--386 (2006)

\bibitem{exploring:topiccoherences}
Stevens, K., Kegelmeyer, P., Andrzejewski, D., Buttler, D.: Exploring topic
  coherence over many models and many topics.
\newblock In: EMNLP-CoNLL, pp. 952--961 (2012)

\bibitem{understand:lda}
Tang, J., Meng, Z., Nguyen, X., Mei, Q., Zhang, M.: Understanding the limiting
  factors of topic modeling via posterior contraction analysis.
\newblock In: ICML, pp. 190--198 (2014)

\bibitem{tong:topicdiscovery}
Tong, Y., Cao, C.C., Chen, L.: Tcs: Efficient topic discovery over
  crowd-oriented service data.
\newblock In: KDD, pp. 861--870. ACM, New York, NY, USA (2014)

\bibitem{wordsim:chinese}
Wang, X., Jia, Y., Zhou, B., Ding, Z., Zheng, L.: Computing semantic
  relatedness using chinese wikipedia links and taxonomy.
\newblock Journal of Chinese Computer Systems \textbf{32}(11), 2237--2242
  (2011)

\bibitem{word:cooccurrence:documentlevel}
Wang, X., McCallum, A.: Topics over time: A non-markov continuous-time model of
  topical trends.
\newblock In: KDD, pp. 424--433 (2006)

\bibitem{twitterrank:twitteraggregation}
Weng, J., Lim, E.P., Jiang, J., He, Q.: Twitterrank: Finding topic-sensitive
  influential twitterers.
\newblock In: WSDM, pp. 261--270 (2010)

\bibitem{shorttext:btm}
Yan, X., Guo, J., Lan, Y., Cheng, X.: A biterm topic model for short texts.
\newblock In: WWW, pp. 1445--1456 (2013)

\bibitem{yu2011trends}
Yu, L., Asur, S., Huberman, B.A.: What trends in chinese social media.
\newblock arXiv preprint arXiv:1107.3522  (2011)

\bibitem{yu2013dynamics}
Yu, L.L., Asur, S., Huberman, B.A.: Dynamics of trends and attention in chinese
  social media.
\newblock arXiv preprint arXiv:1312.0649  (2013)

\bibitem{onetopic:eachtweeter}
Zhao, W.X., Jiang, J., Weng, J., He, J., Lim, E.P., Yan, H., Li, X.: Comparing
  twitter and traditional media using topic models.
\newblock In: ECIR, pp. 338--349 (2011)

\bibitem{kais:querysuggestion}
Zhou, T., Lyu, M.T., King, I., Lou, J.: Learning to suggest questions in social
  media.
\newblock Knowledge and Information Systems pp. 1--28 (2014)

\end{thebibliography}

\end{document}